\pdfoutput=1

\documentclass[11pt]{article}

\usepackage{EACL2023}

\usepackage{times}
\usepackage{latexsym}
\usepackage{graphicx}
\usepackage{makecell}
\usepackage{multirow}
\usepackage{amsmath}
\usepackage{accents}

\usepackage{titlesec} 
\titleformat{\subsubsection}[runin]
  {\normalfont\normalsize\bfseries}{\thesubsubsection}{1em}{}

\usepackage[T1]{fontenc}

\usepackage[utf8]{inputenc}

\usepackage{microtype}

\usepackage{inconsolata}

\title{Enhancing conversational quality in language learning chatbots: \\ An evaluation of GPT4 for ASR error correction}

\author{Long Mai, Julie Carson-Berndsen\\
  ML-Labs, School of Computer Science, University College Dublin, Ireland \\
  \texttt{long.mai@ucdconnect.ie, julie.berndsen@ucd.ie} \\}

\begin{document}
\maketitle
\begin{abstract}
The integration of natural language processing (NLP) technologies into educational applications has shown promising results, particularly in the language learning domain. Recently, many spoken open-domain chatbots have been used as speaking partners, helping language learners improve their language skills. However, one of the significant challenges is the high word-error-rate (WER) when recognizing non-native/non-fluent speech, which interrupts conversation flow and leads to disappointment for learners. This paper explores the use of GPT4 for ASR error correction in conversational settings. In addition to WER, we propose to use semantic textual similarity (STS) and next response sensibility (NRS) metrics to evaluate the impact of error correction models on the quality of the conversation. We find that transcriptions corrected by GPT4 lead to higher conversation quality, despite an increase in WER. GPT4 also outperforms standard error correction methods without the need for in-domain training data.

\end{abstract}

\section{Introduction}

Language learning has always been a complex and challenging endeavour, requiring significant effort and practise to achieve proficiency in a second language. In recent years, advancements in NLP technologies have shown great potential for transforming language learning experiences \cite{nlp_edu}. One particularly promising application of NLP in this domain is the integration of spoken open-domain chatbots as speaking partners for second language acquisition \cite{towards}. The use of chatbots as language learning tools has gained traction due to their ability to provide personalized and interactive conversations, offering learners an opportunity to practise speaking and engage in dialogue. 

\begin{figure}
    \centering
    \includegraphics[width=\linewidth]{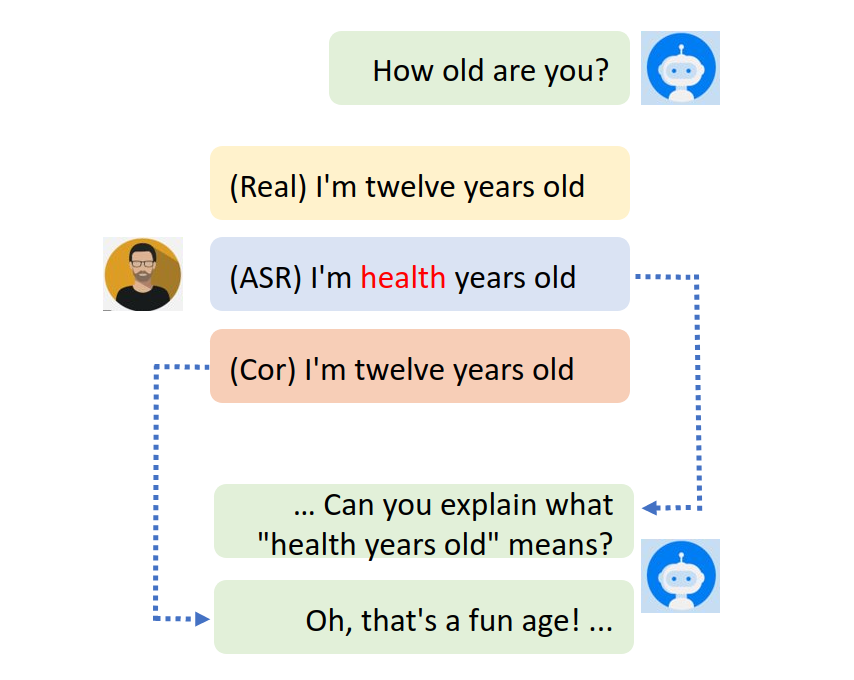}
    \caption{Example conversation between a learner and a chatbot with the presence of ASR errors.}
    \label{fig:cor_exam}
\end{figure}

A spoken open-domain chatbot often contains three main components: (1) an automatic speech recognition (ASR) system transcribing a user's spoken utterances into textual ones, (2) a dialogue response generator (DG) generating relevant responses based on the output of the ASR system, (3) a text to speech (TTS) model transforming bot responses into spoken forms. Because it is a cascade system, errors propagated from the ASR component will lead to unexpected or incorrect bot responses. This is problematic when applying the system to language learners as they frequently make pronunciation, vocabulary, and grammar mistakes that may fail the ASR system \cite{asr_non_native}. Figure \ref{fig:cor_exam} illustrates how ASR errors can lead to confusion and how correction methods can resolve it. \textit{Real} refers to the ground-truth intention of what the learner wants to say, while \textit{Cor} represents the corrected version of the ASR transcription.

This paper tackles the issue of maintaining smooth conversations despite the inherent limitations of ASR systems in transcribing learner responses. We are the first to explore the potential of leveraging GPT4, the most powerful large language model, as an ASR error correction model. More specifically, our contributions can be highlighted as follow:

(1) We propose to use two automatic metrics: semantic textual similarity (STS) and next response sensibility (NRS) as alternatives to WER for measuring the quality of corrected transcriptions. The first metric is utilised to measure the semantic similarity between two transcriptions with respect to dialogue context, whereas the second metric determines whether an ASR transcription can result in sensible subsequent responses from the chatbot. We demonstrate that STS and NRS are in agreement with human judgment and are more effective in evaluating the performance of ASR systems in conversational settings.

(2) We benchmark GPT4 against standard ASR error correction models and observed that it achieves better performances compared to sequence-to-sequence (Seq2Seq) correction methods without requiring domain-specific training data. We found that GPT4 tends to exhibit more over-corrections and miss-corrections, resulting in a higher WER. However, transcriptions corrected by GPT4 demonstrate closer semantic alignment with the real transcription, leading to improved STS and NRS scores. Furthermore, GPT4 surpasses mere word-by-word corrections by modifying grammar and sentence structure to enhance clarity while endeavouring to infer the true intention of the speaker.

Through benchmarking GPT4, we highlight its potential as a valuable tool in language learning. This research contributes to exploring NLP technologies in education, particularly in language learning, addressing challenges of ASR errors.

\section{ASR error correction}

End-to-end models dominate modern speech recognition systems. These are often composed of two components: acoustic modeling and language modeling. Since the language models are often trained on large-scale text-corpora in order to minimize perplexity, they do not take into account the characteristic distribution made by the ASR model \cite{gg-cor}. In other words, the texts used to train the language model might not come from the same distribution as the texts generated by the ASR model during inference, leading to transcription errors.

Previous research addresses errors in transcriptions by attempting to convert incorrect versions into accurate ones. Typically, this involves employing a sequence-to-sequence technique, where the erroneous transcription is used as the input and the corrected version as the output. Consequently, correction models can be developed using architectures like Long Short-Term Memory \cite{gg-cor} and Transformers \cite{nvidia-cor,adapt-cor}. Nonetheless, using solely the 1-best ASR transcription to identify and rectify errors is challenging. Recent research \cite{bart-cor} indicates that only a third of transcription mistakes can be fixed using the 1-best transcription. As such, numerous strategies have been suggested to incorporate supplementary data such as contextual information \cite{uber-cor}, acoustic features \cite{cor_add2}, and visual features \cite{cor_add3}. Standard auto-aggressive correction techniques can be slow and difficult to utilise in a production setting. In response, more recent non-autoregressive methods like FastCorrect \cite{fast-cor} and SoftCorrect \cite{soft-cor} were proposed to expedite the correction process. However, these methods require multiple ASR hypotheses and confidence scores for predicted tokens, which may not always be available in all ASR systems. Moreover, these strategies ignore dialogue context, a crucial factor in error correction in conversational settings.

\section{Methodologies}
\label{sec:method}

\subsection{Seq2Seq error correction} We employ an auto-regressive sequence-to-sequence (Seq2Seq) model for error correction as they show competitive performances to most recent correction models \cite{fast-cor,soft-cor}. More specifically, we use the same multi-input encoder-decoder architecture proposed in \cite{long_mai} to make use of additional information including dialogue context and acoustic features. The model consists of three components: a text encoder, an acoustic encoder, and a decoder. 

The text encoder is a stack of 6 transformer layers, receiving the best ASR transcription and the dialogue context as input. It encodes the text input and produces text output \(H_{t} \in R^{L \times D}\), where $L$ is the number of tokens in the input, and $D$ is the model hidden size.

The input of the acoustic encoder is a sequence of contextualized vectors \(H_{a} \in R^{T \times W}\) obtained by passing the audio through a pre-trained Wav2Vec2-Large-Robust model \cite{wav2vec-robust}, where $T$ is the number of 20ms timesteps and $W$ is the hidden size. A number of VGG convolutional blocks \cite{vgg} and transformer layers are then applied to transform the input acoustic features from \(H_{a} \in R^{T \times W}\) to \(H_{a} \in R^{T/8 \times D}\).

The decoder architecture is nearly identical to the text encoder with 6 transformer layers. The serial combination strategy proposed in \cite{combine} allows the decoder to attend to both text and acoustic encoder outputs.

The training procedure for the model is the same as described in \cite{long_mai}.

\subsection{GPT4 error correction} GPT4, developed by OpenAI\footnote{https://openai.com/}, is a powerful general-purpose chatbot that relies on advanced large language models. Its impressive capabilities, such as code generation, language translation, and mathematical reasoning have garnered significant attention from industry, academia, and the general public. Although the technical specifics of GPT4 have not been comprehensively disclosed, it is known to be built upon the ubiquitous ChatGPT and Instruct-GPT \cite{instruct_gpt}, which undergoes training using instruction tuning and reinforcement learning from human feedback \cite{rlhf}.

Given the dialogue context and ASR transcription, we use GPT4, denoted as GPT4-Cor, as follows:

\textit{You are an expert transcriptionist and have been given the task to review the accuracy of a speech recognition system. Below is a conversation transcription. Identify any speech recognition errors in the transcript and provide the corrected version. Remember, do not correct grammatical errors, sentence structures, verb tenses, repetitions, and hesitations. Your focus should be solely on fixing speech recognition errors.}

\noindent \textit{Conversation Transcription:}

\noindent \textit{Person A:}

\noindent \textit{Person B:}

The line of \textit{Person A} is the chatbot's response, representing the dialogue context with only 1 turn. The line of \textit{Person B} is the ASR transcription of the learner response, which needs to be correct. As GPT4 is trained to generate well-written text, it has a strong tendency to correct all grammar errors in the transcription. However, the goal of ASR error correction is to get the automatic transcriptions to be as close as possible to the real transcriptions, which might contain grammar errors. Therefore, we explicitly told the model to focus solely on fixing speech recognition errors.

\section{Evaluation}
\label{sec:eval}

\subsection{ASR models}
We use the state-of-the-art speech recognition model Whisper-Medium \cite{whisper} to transcribe learner speech. The model is trained on a vast amount of multilingual and multitask supervised data collected from the web. Whisper is known for its robustness and accuracy across different accents and domains, making it an excellent candidate for transcribing non-native/non-fluent speech. To make sure the error correction models are robust, we also experiment with less capable ASR models including Microsoft ASR\footnote{https://azure.microsoft.com/products/cognitive-services/} and NeMo ASR (STT En Conformer-CTC-Large) \cite{nemo}.

\subsection{TLT Corpus: Learner-Bot conversations}
It is best to recruit language learners to engage in multiple-turn conversations with the chatbot and then apply ASR correction methods to the learners' transcriptions to assess if conversation quality improves. However, this practise can be expensive and time-consuming. Additionally, second language users may struggle to communicate effectively with the chatbots, leading to disruptions in the conversation across multiple turns. Given the absence of spoken datasets dedicated to long-term interactions between chatbots and language learners, our study will focus solely on 3-turn conversations. An example of such a conversation is as follows:

\textit{$1^{st}$ turn (Bot): What is your favourite book?}

\textit{$2^{nd}$ turn (Learner): My favourite book is 1984.}

\textit{$3^{rd}$ turn (Bot): That's a great book.}

Our goal is to apply different ASR correction models on the learner transcription at the $2^{nd}$ turn to (1) improve the quality of the transcription itself, and (2) avoid misunderstanding when generating the bot response at the $3^{rd}$ turn.

To simulate these 3-turn conversations, we utilise the TLT-school corpus \cite{tlt}, which is a corpus of non-native children's speech. This corpus consists of speech utterances collected from schools in northern Italy to assess the English language proficiency of students. The recordings were made with students aged between nine and sixteen years, belonging to different CEFR levels, namely A1 (primary school), A2 (secondary school), and B1 (high school). Each student is given a set of questions and is required to provide spoken answers for each question. We transform this data into 3-turn conversations by assigning the question to the $1^{st}$ turn, the student's answer to the $2^{nd}$ turn, and the bot-generated response to the $3^{rd}$ turn. Given that the questions cover various common topics like pets, hobbies, and careers, we consider the corpus to be most suitable for simulating open-domain conversations between language learners and chatbots.

We combine data from TLT-2016, TLT-2017, and TLT-2018, resulting in a total of 51.2 hours of recorded speech. To reduce evaluation efforts, we perform several processing steps on the dataset.

First, we remove long-form questions such as 'Tell me about yourself' or 'Convince someone to do something' as well as situational questions where students were asked to play specific roles like a customer in a restaurant or a library. We also filter out samples with answers that were either too short (less than 2 words) or too long (over 30 seconds or 150 words).

Second, due to the limited vocabulary of the students, many of their answers were highly similar. To avoid doing the repetitive evaluation, we remove highly similar answer pairs if the cosine similarity between them exceeded 0.95. We utilise Sbert-all-MiniLM-L6-v2 \cite{sbert} to compute the similarity score. 

Finally, we eliminate samples where the answers do not address the question or fail to convey any substantial points. This was achieved automatically by prompting GPT4 as follows:

\textit{Check if person B's response has at least a point. Please focus only on the overall meaning of the response. Give a clear answer (Yes or No).}

\noindent \textit{Person A:}

\noindent \textit{Person B:}

Table \ref{tab:tlt} presents the statistics of the TLT corpus before and after processing. We divided the final dataset into two sets: 3203 samples for testing and 2077 samples for training (Seq2Seq correction). We intentionally kept the training set small to mimic a real-world scenario where obtaining a large amount of data specific to the target domain is challenging.

\begin{table}
\centering
\begin{tabular}{llll} 
\hline
& \#Q & \#A   & \#S   \\ 
\hline
No filtering & 84  & 14561 & 3533  \\ 
\hline
\begin{tabular}[c]{@{}l@{}}- Long questions \& \\~ long/short answers\end{tabular} & 57  & 6819  & 2712  \\ 
\hline
\begin{tabular}[c]{@{}l@{}}- Duplicate \& irrelevant\\~ answers\end{tabular}       & 57  & 5279  & 2382  \\
\hline
\end{tabular}
\caption{TLT-Corpus statistics before and after processing. \#Q, \#A, \#S denotes the number of questions, answers, and students respectively.}
\label{tab:tlt}
\end{table}

\subsection{Seq2Seq error correction model}
We train and evaluate the Seq2Seq model with two settings: the in-domain setting, where the model is trained and evaluated on the TLT dataset, and the out-of-domain setting, where the model is trained on a different dataset and then tested on the TLT set. The latter setting is more practical as we often do not know the distribution of the target domain. It is also time-consuming to collect and annotate data for every domain.

In the out-of-domain setting, we train the Seq2Seq model using 967751 correction samples generated from National Speech Corpus (NSC) Part 3 and Part 6 \cite{ncs}. The corpus comprises a total of 2000 hours of speech derived from open-domain conversations between two speakers from Singapore. To the best of our knowledge, NSC is the only large-scale corpus that particularly suits the construction of data for training our ASR correction model because it includes non-native/non-fluent speakers.

In the in-domain setting, we take the pre-trained Seq2Seq model on the NSC corpus and fine-tune it using 2077 samples from the TLT training set.

Each training example is a tuple of (ASR transcription, dialogue context, acoustic features, and real transcription). The ASR transcription is generated using the Whisper ASR model, while the dialogue context represents the concatenated conversation history. Acoustic features are generated using the Wav2Vec2-Large-Robust model \cite{wav2vec-robust} on the audio. The real transcription corresponds to the ground-truth transcription of the audio, annotated by a human.

\section{Evaluation metrics}

\subsection{Semantic textual similarity}
There are limitations to using WER as the sole metric to assess the success of the ASR system, particularly in conversational settings. First, WER only compares the two transcriptions themselves without acknowledging the dialogue context, a crucial information to determine the semantic meaning of the transcription. Second, since the emphasis is on understanding user intention rather than achieving exact word-by-word recognition, and considering the anticipated errors in pronunciation, grammar, and word choices made by language learners, relying solely on WER for transcription comparisons can be sub-optimal.

We propose utilizing semantic textual similarity approaches to evaluate the quality of the automatic transcription with respect to the real transcription. This approach allows us to focus on capturing the semantic meaning and understanding between the two transcriptions rather than solely relying on word matching. The score is calculated as follow: $STS(T,R) = $
\[ 
100 * cos\_sim(Sbert(C+T), Sbert(C+R)))
\]

\begin{table*}
\centering
\begin{tabular}{lll} 
\hline
(Context + \textbf{ASR}, Context + \textbf{Real})  & STS($\uparrow$) & WER($\downarrow$)  \\ 
\hline
\begin{tabular}[c]{@{}l@{}}(What colour are your eyes? [SEP] \textbf{my color ice is blues},\\ \ What colour are your eyes? [SEP] \textbf{my color eyes is blues})\end{tabular} & 91.2 & 20  \\ 
\hline
\begin{tabular}[c]{@{}l@{}}(What colour are your eyes? [SEP] \textbf{my eye color is blue},\\ \ What colour are your eyes? [SEP] \textbf{my color eyes is blues})\end{tabular}  & 94.4 & 60 \\
\hline
\end{tabular}
\caption{Comparing ASR transcriptions with real transcriptions using STS and WER(\%).}
\label{tab:sts}
\end{table*}

We first insert the dialogue context $C$ before each transcription $T$ and $R$. We then employ Sbert-all-MiniLM-L6-v2 \cite{sbert} to encode the concatenated text and use cosine similarity to compute the semantic similarity of the two representations. Table \ref{tab:sts} demonstrates the superiority of using STS over WER when comparing two transcriptions. Note that the WER score is computed based on transcriptions only without dialogue context.

\subsection{Next response sensibility}
\label{sec:nrs}
In the context of spoken dialogue systems, it is crucial to assess the impact of the ASR system on the performance of downstream components like the response generator. We propose to use next response sensibility (NRS) as an indicator of the ASR system's effectiveness in generating transcriptions that contribute to sensible and coherent bot responses.

Given a dialogue context $C$, the real transcription of the speaker's response $R$ and its ASR-generated version $T$. NRS is calculated as follow $NRS(T) = $
\[ 
is\_sensible(C+R, resp\_gen(C+T))
\]

Where $is\_sensible(c,r)$ is a boolean function to check whether the response $r$ is sensible given the dialogue context $c$. The term $resp\_gen(c)$ represents the response generator, which takes into account the previous conversation history $c$ and produces a new response. The notation $C+R$ denotes the concatenation of the response $R$ at the end of the context $C$.

Based on the above calculation of NRS, if the ASR system produces an automatic transcription that is semantically similar to the real transcription, it is highly probable that the subsequent response generated by the chatbot will make sense. Conversely, if the ASR system generates a poor transcription, the resulting next response is likely to be nonsensical. Figure \ref{fig:cor_exam} serves as an example illustrating how NRS is valuable for assessing the quality of an ASR transcription with and without error correction.

\subsubsection*{Response sensibility detector} We utilise GPT4 to perform the task using the following prompt:

\textit{As an expert conversationalist, your task is to evaluate a conversation between two individuals. Assess whether Person A's response is sensible and relevant to Person B's response. Additionally, ensure that Person A does not misunderstand Person B and does not bring up irrelevant entities that were not mentioned by Person B. Finally, check if person A does not ask redundant clarification questions when person B has already provided a valid answer. Provide a clear 'Yes' or 'No' answer indicating whether Person A's response meets all the requirements.}

\noindent \textit{Conversation:}

\noindent \textit{Person A:}

\noindent \textit{Person B:}

\noindent \textit{Person A:}

The first line of \textit{Person A} represents a single-turn dialogue context. The line of \textit{Person B} represents the real transcription of the user, and the last line of \textit{Person A} represents the bot-generated response that needs to be checked for sensibility. This automatic method obtains an accuracy of 86.7\% when compared to 3048 human-annotated samples.

\subsubsection*{Response generation model} 
We use GPT4 as our text-to-text response generation model. Given the dialogue context between two people: \textit{Person A} (chatbot), and \textit{Person B} (learner), we use GPT4, denoted as GPT4-RG, to generate the chatbot's next response as follows:

\textit{You're an expert conversationalist. You will be given a conversation between two people: Person A and Person B. I want you to play the role of Person A and write the next response.}

\noindent \textit{Conversation:}

\noindent \textit{Person A:}

\noindent \textit{Person B:}

We also experiment with other less capable response generation models, such as Blenderbot 3 (BB3) and Blenderbot 1 (BB1), to assess the robustness of the error correction models.

\section{Evaluation results}
This section presents the performance of ASR correction methods using both automatic metrics and human preferences.
\subsubsection*{Evaluation set selection} The evaluation process can be expensive due to the need for numerous API calls with GPT4, as well as manual annotation by human annotators. To reduce this effort, we decide to focus the evaluation on a subset comprising only 'non-easy' samples from the TLT test set. A sample is considered 'easy' when the Whisper ASR system produces a transcription that is clear, sensible, and relevant to the dialogue context, thus requiring no error correction. To identify these 'easy' samples, we prompt GPT4 with the following:

\textit{You're an expert conversationalist. Given a conversation below, determine whether Person B's answer is clear, sensible, relevant to Person A's question. Please provide a clear preference (Yes or No).}

\noindent \textit{Conversation:}

\noindent \textit{Person A:}

\noindent \textit{Person B:}

Where the line of \textit{Person A} represents a single-turn dialogue context, and the line of \textit{Person B} represents the ASR transcription that needs to be checked. By excluding these 'easy' samples, we reduce the number of samples for evaluation from 3203 to just 1016.

\begin{table}[]
\centering
\begin{tabular}{llll}
\hline
Methods & WER & STS & NRS \\ \hline
No correction & 24.6 & 92.9 & 78.1 \\
Seq2Seq (out-domain) & 24.8 & 93.4 & 79.7  \\
Seq2Seq (in-domain) & \textbf{19.6} & 94.7 & 81.5 \\
GPT4-Cor & 33.1 & 94.5 & 82.3 \\
GPT4-Cor (5 shots) & 27.8 & \textbf{95.0} & \textbf{83.2} \\ \hline
\end{tabular}
\caption{Performances of different ASR correction methods in terms of WER(\%), STS, and NRS(\%).}
\label{tab:main}
\end{table}

\subsubsection*{GPT4 Correction} results are shown in Table \ref{tab:main}. Although explicitly instructed to focus on ASR errors, we found that GPT4-Cor still performs grammar error corrections that alter word order and verb tenses, ultimately increasing the WER. This behaviour arises because GPT4-Cor is trained to generate well-written, error-free texts, which may not align perfectly with real transcriptions. However, by providing GPT4-Cor with a few correction examples to guide its approach, the WER can be significantly reduced from 33.1\% to 27.8\%. Overall, GPT4-Cor still contributes a significant improvement in terms of the semantic quality of the transcription, as evidenced by the best increase in NRS from 78.1\% to 83.2\%, surpassing the performance of the Seq2Seq model.

\subsubsection*{Seq2Seq Correction} demonstrates lesser performance to GPT4-Cor in terms of STS and NRS but outperforms it in terms of WER. This is primarily because Seq2Seq approaches are more conservative in their corrections, adhering closely to the original ASR transcription without modifying sentence structures or correcting verb tenses. We also observed that an increase in STS does not always result in an increase in NRS. This can be explained by the fact that if two transcriptions convey the same meaning, but one has more matching words compared to the real transcription, it might receive a higher STS score. This explains why Seq2Seq (in-domain) outperforms GPT4-Cor (zero-shot) in STS but achieves lower performance in terms of NRS.

\begin{table}[]
\centering
\begin{tabular}{l|c}
\hline
\begin{tabular}[c]{@{}l@{}}Dialogue \\ models\end{tabular} & \begin{tabular}[c]{@{}c@{}}NRS with\\ no-correction/correction\end{tabular} \\ \hline
GPT4-RG & 78.1/83.2 \\
BB3 & 54.5/59.7 \\
BB1 & 53.4/58.1 \\ \hline
\end{tabular}
\caption{NRS(\%) with no-correction/correction across different dialogue generation models. Whisper is used for ASR model and GPT4-Cor is used for ASR error correction.}
\label{tab:resp}
\end{table}

\begin{table}[]
\centering
\begin{tabular}{l|ccc}
\hline
\multicolumn{1}{l|}{\multirow{2}{*}{Systems}} & \multicolumn{3}{c}{\begin{tabular}[c]{@{}c@{}}Performances with \\ no-correction/correction\end{tabular}} \\ \cline{2-4} 
\multicolumn{1}{c|}{} & WER & STS & NRS \\ \hline
Whisper & 25/28 & 92.9/95.0 & 78/83 \\
Microsoft & 26/30 & 92.0/94.6 & 78/83 \\
NeMo & 35/38 & 91.3/93.6 & 75/81 \\ \hline
\end{tabular}
\caption{Performances with no-correction/correction across different ASR models. GPT4-Cor is used for ASR error correction. The figures for WER(\%) and NRS(\%) are rounded to the nearest whole number.}
\label{tab:asr}
\end{table}

\subsubsection*{Different response generators} Table \ref{tab:resp} demonstrates that the success in error correction leads to better bot responses across different response generation methods, including GPT4-RG, BB3, and BB1. We expected to observe greater improvements in the NRS score due to error correction for BB1 and BB3 compared to GPT4-RG, as they are less capable and therefore more prone to ASR errors. However, we found that the improvements remained consistent at around 5\% across all models. Upon inspecting the results, we noticed that less capable response models tend to generate more generic responses, such as 'I have not heard about it' or 'I have to check it out'. Fortunately, this type of response helps to avoid the risk of misunderstanding the user. In contrast, GPT4-RG is very specific in its responses, which can sometimes be a disadvantage if there are misrecognized entities in the ASR transcription.

\subsubsection*{Different ASR models} Table \ref{tab:asr} demonstrates the consistent improvements brought by GPT4-Cor  on different ASR models. It is observed that GPT4-Cor achieves greater improvements on less capable ASR models, such as Nemo ASR, with an increase of 2.3 for STS and 6\% for NRS compared to more capable ASR models like Whisper ASR with an increase of 2.1 for STS and 5\% for NRS. For ASR models with comparable performances, GPT4-Cor brings similar improvements.

\begin{table}[]
\centering
\begin{tabular}{lll}
\hline
Methods & SE & NRS \\ \hline
No correction & 56.4 & 68.7 \\
Seq2Seq (in-domain) & 67.2 & 75.3 \\
GPT4-Cor (5 shots) & 71.8 & 78.8 \\ \hline
\end{tabular}
\caption{Human evaluation on different ASR correction methods. SE(\%) stands for semantic equivalent, which measures if two transcriptions convey the same idea.}
\label{tab:human}
\end{table}

\subsubsection*{Human evaluation} To ensure the validity of the improvements brought by ASR error correction and to demonstrate the reliability of STS and NRS as metrics for system comparison, we asked human annotators to assess the quality of Whisper ASR transcriptions and their corrected versions by Seq2Seq and GPT4-Cor. The annotators checked each transcription to determine if it was semantically equivalent to the ground truth transcription, meaning that both responses conveyed the same idea, regardless of hesitations, repetitions, or grammatical errors. They also manually assessed if the corresponding transcription leads to a sensible bot response, as described in Section \ref{sec:nrs}.

As shown in Table \ref{tab:human}, both Seq2Seq and GPT4-Cor bring significant improvements in transcription quality based on human judgment. GPT4-Cor achieves the best results with a 15.4\% increase in SE and a 10.1\% increase in NRS compared to 10.8\% and 7.5\% respectively for Seq2Seq. The results also confirm that STS and NRS (automatic) are better metrics compared to WER since they align with human evaluation scores. We expected the numbers for SE and NRS to be close as if an automatic transcription is semantically equivalent to the real transcription, the bot response should be sensible. However, we observed that the figure for NRS is much higher. This can be attributed to the fact that GPT4-RG sometimes manages to infer the true intention of the speaker and generate a sensible response, even when the automatic transcription and the real transcription differ significantly in terms of semantics.

\section{Correction examples}

\begin{table*}[]
\centering
\begin{tabular}{l|ll}
\hline
Id & \multicolumn{1}{c}{Questions} & \multicolumn{1}{c}{Learner responses} \\ \hline
\multirow{3}{*}{1} & \multirow{3}{*}{\begin{tabular}[c]{@{}l@{}}Was it your first time in \\ York?\end{tabular}} & (R) I go in York I go last  in York  at November \\
 &  & (A) I go last in York at November \\
 &  & (C) I went last to York in November \\ \hline
\multirow{3}{*}{2} & \multirow{3}{*}{\begin{tabular}[c]{@{}l@{}}Do you prefer outdoor\\ or indoor sports?\end{tabular}} & \begin{tabular}[c]{@{}l@{}}(R) I prefer indoor sport because I play I play ski and there is\\  snowboarding...\end{tabular} \\
 &  & (A) I prefer windows port because I play  ski and snowboarding \\
 &  & (C) I prefer outdoor sports because I play ski and snowboarding \\ \hline
 \multirow{3}{*}{3} & \multirow{3}{*}{\begin{tabular}[c]{@{}l@{}}Do you want something to\\ drink?\end{tabular}} & (R) Thank you I want an coke \\
 &  & (A) Thank you I want to cook \\
 &  & (C) Thank you I want a coke \\ \hline
\multirow{3}{*}{4} & \multirow{3}{*}{What animals do you like?} & (R) I like a cat \\
 &  & \begin{tabular}[c]{@{}l@{}}(A) I like rocket\\ → GPT4-RG: Oh, I think you might be talking about a \\ character named Rocket from Guardians of the Galaxy...\end{tabular} \\
 &  & \begin{tabular}[c]{@{}l@{}}(C) I like raccoons\\ → GPT4-RG: Oh, raccoons are cute and intelligent \\ creatures. What is it about raccoons that you like the most?\end{tabular} \\ \hline
\end{tabular}
\caption{Correction samples generated by GPT4. (R), (A), (C) refer to the real, automatic, and corrected transcriptions respectively.}
\label{tab:exam}
\end{table*}

After analyzing the corrected transcriptions provided by GPT4-Cor and presenting a selection of representative samples in Table \ref{tab:exam}, we have gained valuable insights:

\subsubsection*{Grammar and structure corrections} Example 1 demonstrates that GPT4-Cor not only focuses on ASR errors but also addresses grammar and sentence structure issues in the transcriptions. This leads to an increase in WER but also yields a more coherent and natural-sounding transcription, improving the overall intelligibility of the conversation.

\subsubsection*{Inference of speaker intent} One of the most notable findings is that GPT4-Cor appears capable of inferring the speaker's true intent even when the real transcription contains logical errors. We have observed that learners' speech do not always align precisely with their intended meaning, as they often make pronunciation mistakes or use incorrect vocabulary. In example 2, despite the learner saying, 'I prefer indoor sport because I play I play ski and there is snowboarding', this statement is logically flawed since skiing and snowboarding are commonly outdoor sports. However, based on the context, GPT4-Cor can infer that the learner's true preference is outdoor sports.

\subsubsection*{Utilization of dialogue context} As demonstrated in example 3, the ASR makes a mistake by transcribing the learner's response as 'I want to cook', which is illogical given the dialogue context of 'Do you want something to drink?'. Nevertheless, GPT4-Cor, equipped with the dialogue context, can identify the issue, perform logical reasoning, and utilise phonetic search to rectify the erroneous ASR transcription to 'I want a coke'. This ability to leverage dialogue context serves as a valuable complement to many context-independent ASR systems.

\subsubsection*{Miss-corrections} In the final example 4, it can be observed that GPT4 makes an error by providing the corrected transcription as 'I like raccoons'. This occurs because the ASR transcription 'I like rocket' differs significantly phonetically from the ground truth 'I like a cat'. However, the miss-correction does not worsen the outcome compared to having no correction, as both versions result in the generation of bot responses that are not sensible, requiring the learner to repeat the answer.

\section{Conclusion}

This paper investigates the use of GPT4 for ASR error correction in conversational settings, focusing on language learning applications. We introduce two metrics, STS and NRS, to evaluate the impact of error correction on conversation quality. The results demonstrate that transcriptions corrected by GPT4 lead to higher conversation quality, despite an increase in WER. GPT4 also outperforms standard error correction methods without the need for in-domain training data.

\section{Acknowledgement}
This work was funded by Science Foundation Ireland through the SFI Centre for Research Training in Machine Learning (18/CRT/6183).

\bibliography{custom}

\begin{thebibliography}{23}
\expandafter\ifx\csname natexlab\endcsname\relax\def\natexlab#1{#1}\fi

\bibitem[{Bibauw et~al.(2019)Bibauw, Fran{\c{c}}ois, and Desmet}]{nlp_edu}
Serge Bibauw, Thomas Fran{\c{c}}ois, and Piet Desmet. 2019.
\newblock Discussing with a computer to practice a foreign language: Research
  synthesis and conceptual framework of dialogue-based call.
\newblock \emph{Computer Assisted Language Learning}, 32(8):827--877.

\bibitem[{Christiano et~al.(2017)Christiano, Leike, Brown, Martic, Legg, and
  Amodei}]{rlhf}
Paul~F Christiano, Jan Leike, Tom Brown, Miljan Martic, Shane Legg, and Dario
  Amodei. 2017.
\newblock Deep reinforcement learning from human preferences.
\newblock \emph{Advances in neural information processing systems}, 30.

\bibitem[{Du et~al.(2022)Du, Pu, Dong, Jin, Qi, Gu, Wu, and Zhou}]{cor_add2}
Jing Du, Shiliang Pu, Qinbo Dong, Chao Jin, Xin Qi, Dian Gu, Ru~Wu, and Hongwei
  Zhou. 2022.
\newblock Cross-modal asr post-processing system for error correction and
  utterance rejection.
\newblock \emph{arXiv preprint arXiv:2201.03313}.

\bibitem[{Gretter et~al.(2020)Gretter, Matassoni, Bann{\`o}, and
  Falavigna}]{tlt}
Roberto Gretter, Marco Matassoni, Stefano Bann{\`o}, and Daniele Falavigna.
  2020.
\newblock Tlt-school: a corpus of non native children speech.
\newblock \emph{arXiv preprint arXiv:2001.08051}.

\bibitem[{Guo et~al.(2019)Guo, Sainath, and Weiss}]{gg-cor}
Jinxi Guo, Tara~N Sainath, and Ron~J Weiss. 2019.
\newblock A spelling correction model for end-to-end speech recognition.
\newblock In \emph{ICASSP 2019-2019 IEEE International Conference on Acoustics,
  Speech and Signal Processing (ICASSP)}, pages 5651--5655. IEEE.

\bibitem[{Hrinchuk et~al.(2020)Hrinchuk, Popova, and Ginsburg}]{nvidia-cor}
Oleksii Hrinchuk, Mariya Popova, and Boris Ginsburg. 2020.
\newblock Correction of automatic speech recognition with transformer
  sequence-to-sequence model.
\newblock In \emph{ICASSP 2020-2020 IEEE International Conference on Acoustics,
  Speech and Signal Processing (ICASSP)}, pages 7074--7078. IEEE.

\bibitem[{Hsu et~al.(2021)Hsu, Sriram, Baevski, Likhomanenko, Xu, Pratap, Kahn,
  Lee, Collobert, Synnaeve et~al.}]{wav2vec-robust}
Wei-Ning Hsu, Anuroop Sriram, Alexei Baevski, Tatiana Likhomanenko, Qiantong
  Xu, Vineel Pratap, Jacob Kahn, Ann Lee, Ronan Collobert, Gabriel Synnaeve,
  et~al. 2021.
\newblock Robust wav2vec 2.0: Analyzing domain shift in self-supervised
  pre-training.
\newblock \emph{arXiv preprint arXiv:2104.01027}.

\bibitem[{Koh et~al.(2019)Koh, Mislan, Khoo, Ang, Ang, Ng, and Tan}]{ncs}
Jia~Xin Koh, Aqilah Mislan, Kevin Khoo, Brian Ang, Wilson Ang, Charmaine Ng,
  and YY~Tan. 2019.
\newblock Building the singapore english national speech corpus.
\newblock \emph{Malay}, 20(25.0):19--3.

\bibitem[{Kuchaiev et~al.(2019)Kuchaiev, Li, Nguyen, Hrinchuk, Leary, Ginsburg,
  Kriman, Beliaev, Lavrukhin, Cook et~al.}]{nemo}
Oleksii Kuchaiev, Jason Li, Huyen Nguyen, Oleksii Hrinchuk, Ryan Leary, Boris
  Ginsburg, Samuel Kriman, Stanislav Beliaev, Vitaly Lavrukhin, Jack Cook,
  et~al. 2019.
\newblock Nemo: a toolkit for building ai applications using neural modules.
\newblock \emph{arXiv preprint arXiv:1909.09577}.

\bibitem[{Leng et~al.(2023)Leng, Tan, Liu, Song, Wang, Li, Qin, Lin, and
  Liu}]{soft-cor}
Yichong Leng, Xu~Tan, Wenjie Liu, Kaitao Song, Rui Wang, Xiang-Yang Li, Tao
  Qin, Ed~Lin, and Tie-Yan Liu. 2023.
\newblock Softcorrect: Error correction with soft detection for automatic
  speech recognition.
\newblock In \emph{Proceedings of the AAAI Conference on Artificial
  Intelligence}, volume~37, pages 13034--13042.

\bibitem[{Leng et~al.(2021)Leng, Tan, Wang, Zhu, Xu, Liu, Liu, Qin, Li, Lin
  et~al.}]{fast-cor}
Yichong Leng, Xu~Tan, Rui Wang, Linchen Zhu, Jin Xu, Wenjie Liu, Linquan Liu,
  Tao Qin, Xiang-Yang Li, Edward Lin, et~al. 2021.
\newblock Fastcorrect 2: Fast error correction on multiple candidates for
  automatic speech recognition.
\newblock \emph{arXiv preprint arXiv:2109.14420}.

\bibitem[{Libovick{\`y} et~al.(2018)Libovick{\`y}, Helcl, and
  Mare{\v{c}}ek}]{combine}
Jind{\v{r}}ich Libovick{\`y}, Jind{\v{r}}ich Helcl, and David Mare{\v{c}}ek.
  2018.
\newblock Input combination strategies for multi-source transformer decoder.
\newblock \emph{arXiv preprint arXiv:1811.04716}.

\bibitem[{Mai and Carson-Berndsen(2022)}]{long_mai}
Long Mai and Julie Carson-Berndsen. 2022.
\newblock Unsupervised domain adaptation for speech recognition with
  unsupervised error correction.
\newblock \emph{arXiv preprint arXiv:2209.12043}.

\bibitem[{Mani et~al.(2020)Mani, Palaskar, Meripo, Konam, and
  Metze}]{adapt-cor}
Anirudh Mani, Shruti Palaskar, Nimshi~Venkat Meripo, Sandeep Konam, and Florian
  Metze. 2020.
\newblock Asr error correction and domain adaptation using machine translation.
\newblock In \emph{ICASSP 2020-2020 IEEE International Conference on Acoustics,
  Speech and Signal Processing (ICASSP)}, pages 6344--6348. IEEE.

\bibitem[{Ouyang et~al.(2022)Ouyang, Wu, Jiang, Almeida, Wainwright, Mishkin,
  Zhang, Agarwal, Slama, Ray et~al.}]{instruct_gpt}
Long Ouyang, Jeffrey Wu, Xu~Jiang, Diogo Almeida, Carroll Wainwright, Pamela
  Mishkin, Chong Zhang, Sandhini Agarwal, Katarina Slama, Alex Ray, et~al.
  2022.
\newblock Training language models to follow instructions with human feedback.
\newblock \emph{Advances in Neural Information Processing Systems},
  35:27730--27744.

\bibitem[{Radford et~al.(2022)Radford, Kim, Xu, Brockman, McLeavey, and
  Sutskever}]{whisper}
Alec Radford, Jong~Wook Kim, Tao Xu, Greg Brockman, Christine McLeavey, and
  Ilya Sutskever. 2022.
\newblock Robust speech recognition via large-scale weak supervision.
\newblock \emph{arXiv preprint arXiv:2212.04356}.

\bibitem[{Reimers and Gurevych(2019)}]{sbert}
Nils Reimers and Iryna Gurevych. 2019.
\newblock Sentence-bert: Sentence embeddings using siamese bert-networks.
\newblock \emph{arXiv preprint arXiv:1908.10084}.

\bibitem[{Simonyan and Zisserman(2014)}]{vgg}
Karen Simonyan and Andrew Zisserman. 2014.
\newblock Very deep convolutional networks for large-scale image recognition.
\newblock \emph{arXiv preprint arXiv:1409.1556}.

\bibitem[{Tyen et~al.(2022)Tyen, Brenchley, Caines, and Buttery}]{towards}
Gladys Tyen, Mark Brenchley, Andrew Caines, and Paula Buttery. 2022.
\newblock \href {https://doi.org/10.18653/v1/2022.bea-1.28} {Towards an
  open-domain chatbot for language practice}.
\newblock In \emph{Proceedings of the 17th Workshop on Innovative Use of NLP
  for Building Educational Applications (BEA 2022)}, pages 234--249, Seattle,
  Washington. Association for Computational Linguistics.

\bibitem[{Weng et~al.(2020)Weng, Miryala, Khatri, Wang, Zheng, Molino,
  Namazifar, Papangelis, Williams, Bell et~al.}]{uber-cor}
Yue Weng, Sai~Sumanth Miryala, Chandra Khatri, Runze Wang, Huaixiu Zheng, Piero
  Molino, Mahdi Namazifar, Alexandros Papangelis, Hugh Williams, Franziska
  Bell, et~al. 2020.
\newblock Joint contextual modeling for asr correction and language
  understanding.
\newblock In \emph{ICASSP 2020-2020 IEEE International Conference on Acoustics,
  Speech and Signal Processing (ICASSP)}, pages 6349--6353. IEEE.

\bibitem[{Wills et~al.(2023)Wills, Bai, Tejedor-Garcia, Cucchiarini, and
  Strik}]{asr_non_native}
Simone Wills, Yu~Bai, Cristian Tejedor-Garcia, Catia Cucchiarini, and Helmer
  Strik. 2023.
\newblock Automatic speech recognition of non-native child speech for language
  learning applications.
\newblock \emph{arXiv preprint arXiv:2306.16710}.

\bibitem[{Xu et~al.(2021)Xu, Li, Zhou, Li, Wang, Cao, Huang, and
  Mao}]{cor_add3}
Heng-Da Xu, Zhongli Li, Qingyu Zhou, Chao Li, Zizhen Wang, Yunbo Cao, Heyan
  Huang, and Xian-Ling Mao. 2021.
\newblock Read, listen, and see: Leveraging multimodal information helps
  chinese spell checking.
\newblock \emph{arXiv preprint arXiv:2105.12306}.

\bibitem[{Zhao et~al.(2021)Zhao, Yang, Wang, Gao, Yan, and Zhou}]{bart-cor}
Yun Zhao, Xuerui Yang, Jinchao Wang, Yongyu Gao, Chao Yan, and Yuanfu Zhou.
  2021.
\newblock Bart based semantic correction for mandarin automatic speech
  recognition system.
\newblock \emph{arXiv preprint arXiv:2104.05507}.

\end{thebibliography}
\bibliographystyle{acl_natbib}

\appendix

\end{document}